\documentclass[runningheads]{llncs}
\usepackage[T1]{fontenc}
\usepackage{appendix}
\usepackage{graphicx}
\usepackage{amsmath}
\usepackage{enumitem}
\usepackage{booktabs}
\usepackage{array}
\usepackage[most]{tcolorbox}
\usepackage{listings}
\usepackage{xcolor}
\usepackage{multirow}

\newtcolorbox{prompt}[2][]{
    enhanced,
    colback=white,
    colframe=black!70,
    boxrule=0.5pt,
    arc=1.5mm,
    fontupper=\small,
    title={#2}, 
    fonttitle=\bfseries\footnotesize,
    coltitle=black,
    colbacktitle=gray!10, 
    attach boxed title to top left={xshift=3mm, yshift=-2mm},
    boxed title style={size=small, boxrule=0.5pt, colframe=black!70},
    top=4mm,
    #1 
}

\lstdefinelanguage{json}{
    basicstyle=\normalfont\ttfamily\small,
    numbers=left,
    numberstyle=\scriptsize,
    stepnumber=1,
    numbersep=8pt,
    showstringspaces=false,
    breaklines=true,         
    breakatwhitespace=false,
    frame=lines,
    backgroundcolor=\color{gray!5},
    commentstyle=\color{gray},
    stringstyle=\color{blue},
    keywordstyle=\color{magenta},
    morekeywords={}, 
    literate=
     *{:}{{{\color{red}{:}}}}{1}
      {,}{{{\color{red}{,}}}}{1}
      {\{}{{{\color{blue}{\{}}}}{1}
      {\}}{{{\color{blue}{\}}}}}{1}
      {[}{{{\color{blue}{[}}}}{1}
      {]}{{{\color{blue}{]}}}}{1},
}

\makeatletter
\renewcommand\subsubsection{\@startsection{subsubsection}{3}{\z@}%
                {-3.25ex\@plus -1ex \@minus -.2ex}
                {-0.5em}
                {\normalfont\normalsize\itshape}} 
\makeatother

\begin{document}
\title{SoccerRef-Agents: Multi-Agent System for Automated Soccer Refereeing}
%
%
\author{Zi Meng\inst{1,2} \and
Wanli Song\inst{1} \and
Yi Hu\inst{2} \and
Jiayuan Rao\textsuperscript{$\dagger$} \inst{2} \and
Gang Chen\textsuperscript{$\dagger$}\inst{2}}
\authorrunning{Z. Meng et al.}
%
\institute{University of Michigan, Michigan, USA\\
\email{\{mengzi,wanlis\}@umich.edu}\\
\and
Shanghai Jiao Tong University, Shanghai, China\\
\email{\{huyi\_0811, jy\_rao, chengang76\}@sjtu.edu.cn}
}
\maketitle              
{\let\thefootnote\relax\footnotetext{\textsuperscript{$\dagger$} Corresponding author.}}
\begin{abstract} Refereeing is vital in sports, where fair, accurate, and explainable decisions are fundamental. While intelligent assistant technologies are being widely adopted in soccer refereeing, current AI-assisted approaches remain preliminary. Existing research mostly focuses on isolated video perception tasks and lacks the ability to understand and reason about foul scenarios. To fill this gap, we propose \textbf{SoccerRef-Agents}, a holistic and explainable multi-agent decision-making framework for soccer refereeing. The main contributions are: (i) constructing the multimodal benchmark \textbf{SoccerRefBench} with over 1,200 referee theory questions and 600 foul video clips; (ii) building a vector-based knowledge base \textbf{RefKnowledgeDB} using the latest \textit{“Laws of the Game”} and a classic case database for precise, knowledge-driven reasoning; (iii) designing a novel multi-agent architecture that collaborates via cross-modal RAG to bridge the semantic gap between visual content and regulatory texts. This work explores the technical capability of integrating MLLMs with refereeing expertise, and evaluations show our system significantly outperforms general-purpose MLLMs in decision accuracy and explanation quality. All databases, benchmarks, and code will be made available.
\end{abstract}

\section{Introduction}\label{Sec: Introduction}

\begin{figure}[t]
    \centering
    \includegraphics[width=1.0\linewidth]{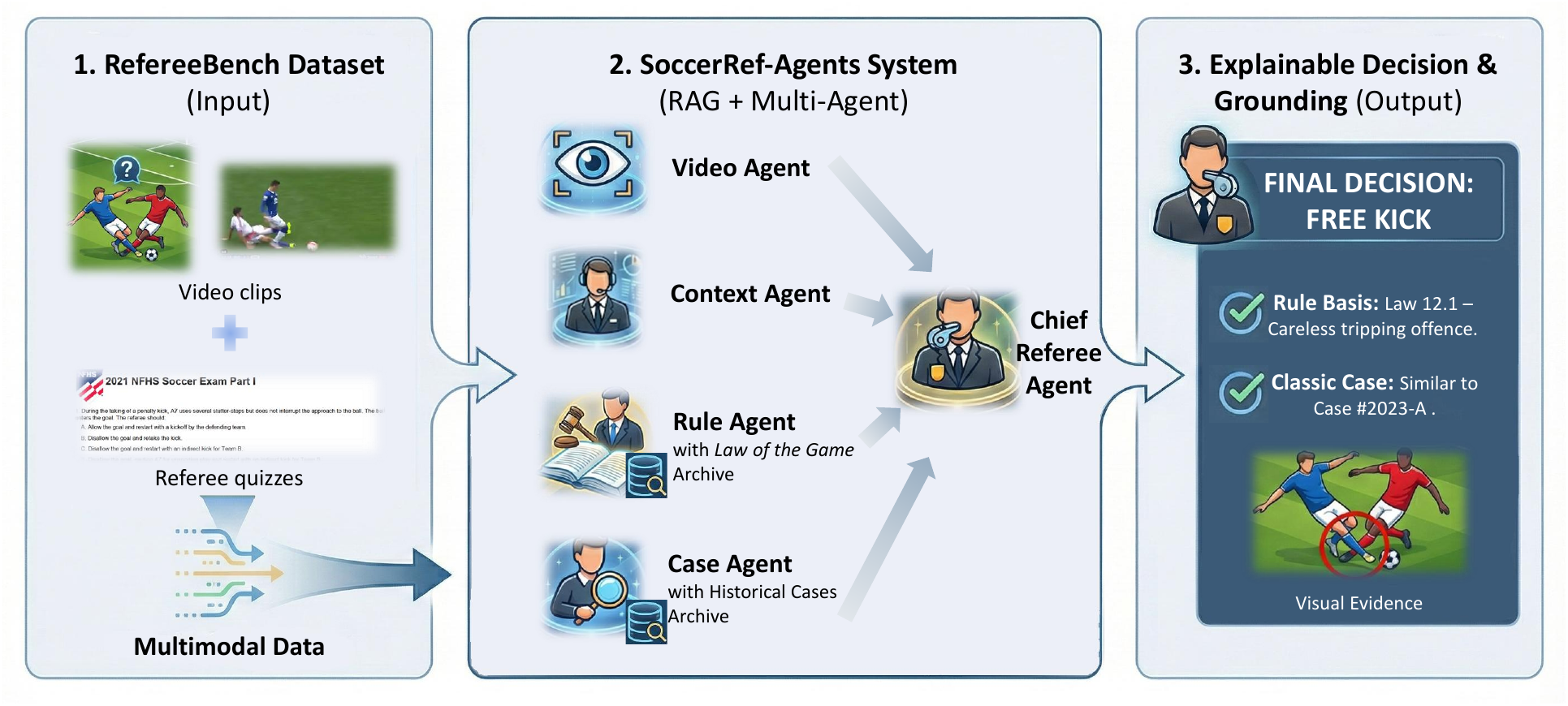}
    \caption{Overview of \textbf{SoccerRef-Agents}. The system mimics a professional officiating team by decomposing the task into perception (Video Agent), background analysis (Context Agent), legal interpretation (Rule Agent), and precedent retrieval (Case Agent), culminating in a final decision by the Chief Referee Agent.}
    \label{fig:teaser}
\end{figure}

Competitive sports captivate global audiences with their inherent dynamism and passion. Among them, soccer, hailed as the most beautiful game, enjoys unparalleled worldwide attention. In modern soccer, with advancements in science and technology, particularly the recent development of AI, technology is being applied across all aspects from training and competition to broadcasting. Within this context, soccer refereeing occupies a special position, as it is tasked with the crucial work of maintaining the orderly conduct of matches. The fairness and integrity of a match are highly dependent on the accuracy of refereeing decisions. Currently, the latest advancements in artificial intelligence have been applied to many aspects of soccer understanding, including video comprehension \cite{comment_gen1,action_recognition1,action_recognition2,action_recognition3,action_recognition4,matchtime} However, in the domain of automated refereeing, despite some attempts, three main challenges remain: \textbf{(i) The Gap Between Perception and Reasoning:} Most existing systems\cite{VARS,X-Vars} treat foul detection as a simple visual classification task. However, effective officiating requires a cognitive leap from visual data to rule-based logic. For instance, determining disciplinary actions such as distinguishing between "reckless" and "using excessive force" is a knowledge-intensive process that necessitates explicit reference to the \textit{"Laws of the Game(LOTG)"}\cite{IFAB_LOTG_2025}, a capability largely absent in current vision-only models. \textbf{(ii) Unreliable and Unexplainable Outputs:} General-purpose multimodal large models often struggle to provide reliable justifications for their decisions. Due to a lack of internalized domain knowledge, these models\cite{hallucination1,hallucination2,MLLM} may generate seemingly plausible but factually incorrect explanations, known as hallucinations. They typically fail to ground their judgments in specific rules or similar historical cases, limiting their practical utility as transparent decision-support systems. \textbf{(iii) Limitations in Sample Size:} Most current researches\cite{sample_size1,sample_size2}  focus on analyzing and processing small-scale samples, lacking a truly large-scale benchmark for the comprehensive evaluation of AI refereeing systems.

To bridge this gap, we propose a comprehensive framework for holistic and explainable soccer refereeing. Recognizing that a valid judgment requires both accurate perception and authoritative knowledge, we first construct \textbf{SoccerRefBench}, a multimodal benchmark specifically designed to evaluate refereeing logic. Unlike general sports datasets, \textbf{SoccerRefBench} integrates theoretical knowledge derived from professional certification exams with practical judgment from controversial broadcast replay clips. Furthermore, to enable knowledge-driven reasoning, we construct \textbf{RefKnowledgeDB} by digitizing and vectorizing the latest \textit{Laws of the Game}\cite{IFAB_LOTG_2025} and a curated database of classic historical cases, serving as the long-term memory of our system.

To tackle the complexity of decision-making, we introduce \textbf{SoccerRef-Agents}, a novel multi-agent system that mimics the collaborative workflow of a professional officiating team. As illustrated in Figure \ref{fig:teaser}, our system decomposes the refereeing task into specialized roles: a \textit{Video Agent} for perception, a \textit{Rule Agent} for legal interpretation, a \textit{Case Agent} for precedent retrieval, and a \textit{Chief Referee Agent} for final adjudication. By leveraging a cross-modal Retrieval-Augmented Generation (RAG) mechanism, our agents can consult the rulebook based on visual analysis, ensuring decisions are not only accurate but also legally cited.

Concretely, we make the following contributions in this paper:
\begin{itemize}
    \item[(i)] We construct \textbf{SoccerRefBench}, a specialized multimodal benchmark for soccer officiating. It comprises 1,218 theoretical exam questions and 600 annotated video clips of fouls, mapped to standard disciplinary outcomes (No Offence, Normal, Yellow, Red).
    \item[(ii)] We establish a specialized vector knowledge base: the \textbf{RefKnowledgeDB}. The database enables precise and fine-grained retrieval of regulations and historical precedents, mitigating model hallucinations.
    \item[(iii)] We introduce \textbf{SoccerRef-Agents}, a multi-agent framework featuring a novel cross-modal reasoning pipeline. By utilizing video descriptions to drive textual knowledge retrieval, our system effectively bridges the semantic gap between visual footage and legal texts.
    \item[(iv)] Extensive evaluations on \textbf{SoccerRefBench} demonstrate the superiority of our agentic system. \textbf{SoccerRef-Agents} significantly outperforms general-purpose MLLMs in decision accuracy and achieves state-of-the-art performance in generating legally grounded explanations.
\end{itemize}

\section{Related Works}\label{Sec: Related Works}
\subsection{Sports Understanding}\label{Subsec: Sports Understanding}
Sports understanding has emerged as a pivotal testbed for evaluating Multimodal Large Language Models (MLLMs) due to its dynamic nature and complex rules. Early research primarily focused on atomic tasks such as action recognition\cite{action_recognition1,action_recognition2,action_recognition3,action_recognition4} and automated scoring\cite{automated_scoring1,automated_scoring2,automated_scoring3}. With the advent of large-scale benchmarks\cite{large_scale_benchmark1,large_scale_benchmark2,large_scale_benchmark3}, the focus has shifted towards holistic understanding. For instance, Sports-QA\cite{SportsQA} and SPORTU\cite{Sportu}  evaluate models on QA tasks across multiple sports disciplines, requiring agents to perceive actions and reason about game states. Recent works have also explored dense video captioning\cite{SoccerNet_Caption,video_caption1} and commentary generation\cite{comment_gen1,matchtime} to enhance fan engagement. However, these general benchmarks often treat sports rules as implicit background knowledge rather than explicit logic constraints, limiting their ability to evaluate fine-grained professional judgment.
\subsection{Soccer Understanding}\label{Subsec: Soccer Understanding}
As the world's most popular sport, soccer has attracted significant research attention. The field is largely driven by extensive datasets like SoccerNet\cite{SoccerNet} , which facilitates tasks ranging from action spotting to replay grounding. Beyond perception, recent studies have ventured into higher-level cognitive tasks. MatchTime\cite{matchtime} and SoccerNet-Caption\cite{SoccerNet_Caption} focus on generating time-aligned commentary, while other works explore tactical analysis\cite{soccer_tactic1,soccer_tactic2} and game state reconstruction\cite{game_reconstruction}. despite these advancements, existing soccer understanding models predominantly focus on descriptive tasks\cite{soccer_description1,soccer_description2} or statistical analysis\cite{soccer_statistic1,soccer_statistic2,soccer_statistic3}. They rarely address the normative aspect of the game—specifically, judging player actions against the strict, textual framework of the Laws of the Game\cite{IFAB_LOTG_2025}.

\subsection{Soccer Refereeing}\label{Subsec: AI Refereeing and Rule Interpretation}
The automation of refereeing has advanced significantly in both industrial applications and academic research, albeit with distinct focuses. In professional leagues, hardware-centric systems like Video Assistant Referee (VAR) \cite{VAR} and Semi-Automated Offside Technology (SAOT) \cite{FIFA_SAOT_2023} provide precise spatial measurements. However, they lack the semantic reasoning capabilities required to judge subjective fouls or interpret player intent. Conversely, academic research focuses on vision-based foul recognition from broadcast footage. For instance, VARS \cite{VARS} treats refereeing as a multi-view classification problem, while X-VARS \cite{X-Vars} explores utilizing MLLMs to generate descriptive explanations for foul events.

Despite these strides, a fully autonomous AI Referee remains elusive. Recent trends, such as the SoccerNet Challenges \cite{soccernet_challenge2024,soccernet_challenge2025}, predominantly frame officiating as a video classification task, mapping visual features directly to disciplinary labels based on statistical correlations. However, authentic refereeing is a complex legal adjudication process, not merely a perceptual one. Existing models \cite{soccernet_challenge2024,soccernet_challenge2025,foul_classification} fail to logically derive verdicts by verifying factual preconditions and aligning them with specific regulatory clauses. This disconnect between visual perception and legal reasoning limits the reliability and explainability of current approaches. To bridge this gap, effective AI systems must move beyond pattern recognition to knowledge-driven reasoning. However, existing datasets lack the explicit alignment between model intuition and the strict textual framework of the \textit{Laws of the Game}\cite{IFAB_LOTG_2025} required to develop and evaluate this capability. This necessity motivates the construction of \textbf{SoccerRefBench}, establishing the foundational resources detailed in Sec.~\ref{Sec: Dataset Construction}

\section{Dataset Construction}\label{Sec: Dataset Construction}
To bridge the critical gap between model intuition and legal adjudication in automated officiating, we introduce a comprehensive data framework designed for knowledge-driven reasoning. This section first outlines the motivation and overview of our proposed framework in Sec.~\ref{subsec: Motivation-Overview}. We then provide detailed descriptions of the data collection and curation processes for our multimodal benchmark, \textbf{SoccerRefBench}, in Sec.~\ref{subsec: Data Collection} and Sec.~\ref{Subsec: Data Curation}, respectively. Finally, we elaborate on the construction of our specialized domain repository, the \textbf{RefKnowledgeDB}, in Sec.~\ref{Subsec: Knowledge Base Construction for RAG}, which serves as the foundation for the system's retrieval capabilities.

\subsection{Motivation \& Overview}\label{subsec: Motivation-Overview}
Soccer refereeing is a highly specialized domain that requires precise interpretation of the rule and split-second decision-making. While existing benchmarks primarily focus on action recognition or general sports question-answering, they still lack evaluation standards specifically tailored for soccer refereeing, particularly in multimodal understanding and rule-based reasoning. To provide an evaluation platform for such a professional scenario, we introduce \textbf{SoccerRefBench}, a comprehensive multimodal benchmark designed to assess automated officiating according to professional refereeing standards, comprises two distinct modalities: (i) a textual subset assessing theoretical knowledge via standardized exams, and (ii) a video subset for evaluating practical judgment in controversial foul scenarios. To enable better reasoning and evidence retrieval in the automated officiating pipeline, we construct a searchable knowledge base \textbf{RefKnowledgeDB}, containing the official \textit{Laws of the Game}\cite{IFAB_LOTG_2025} and historical precedents, thereby supporting knowledge-driven reasoning.

\subsection{Data Collection}\label{subsec: Data Collection}
\begin{figure}[t]
    \centering
    \includegraphics[width=1.0\linewidth]{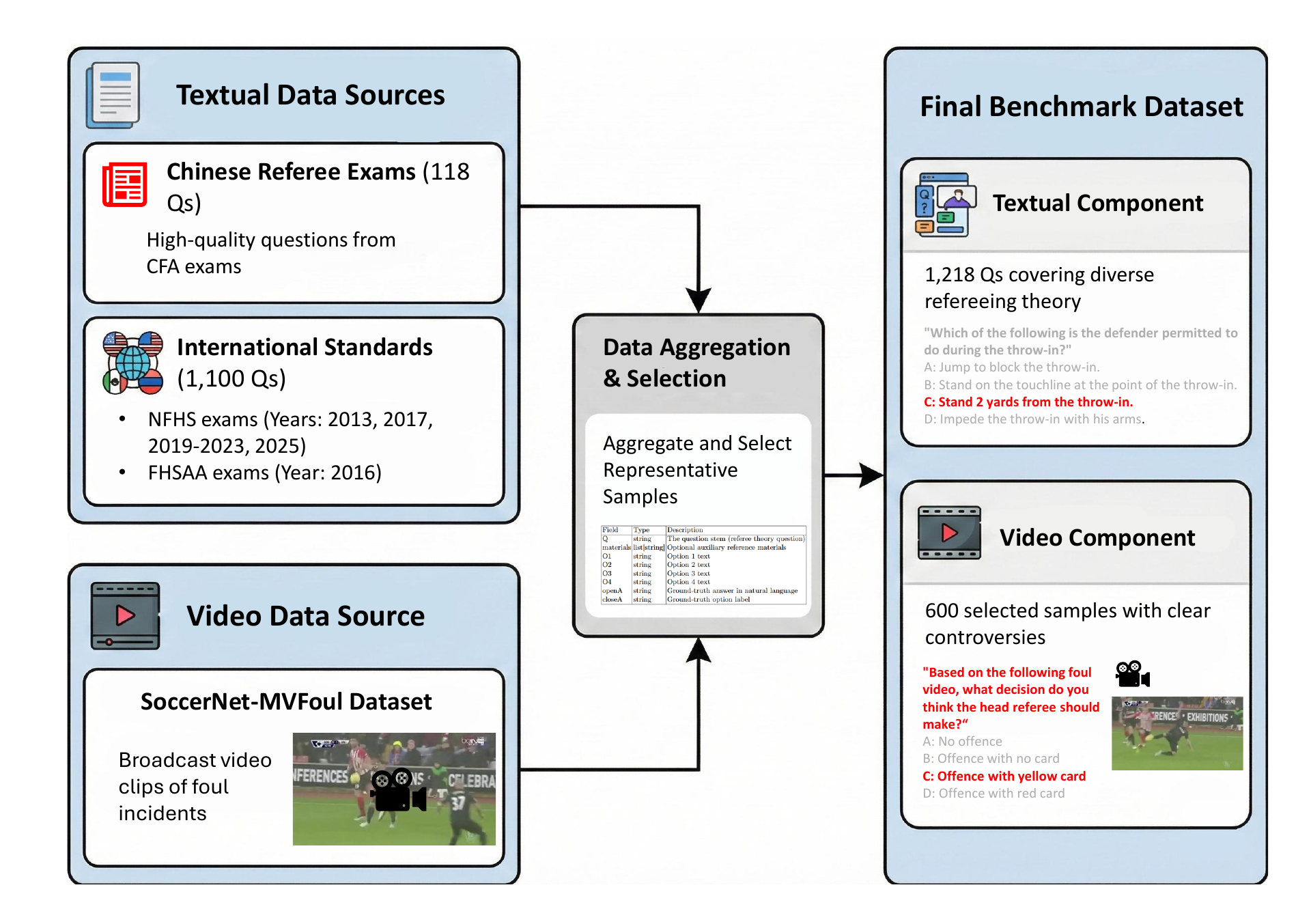}
    \caption{Overview of the data collection pipeline for \textbf{SoccerRefBench}. The dataset integrates 1,218 textual theory questions from diverse international sources and 600 video-based judgment scenarios derived from the SoccerNet-MVFoul dataset.}
    \label{fig:data_collection_pipeline}
\end{figure}

To construct a comprehensive evaluation benchmark, we aggregate data from authoritative refereeing examinations and large-scale video datasets. The overall data collection and aggregation pipeline is illustrated in Figure \ref{fig:data_collection_pipeline}.

\subsubsection{Textual Data Collection.} We collect a total of 1,218 multiple-choice questions covering diverse aspects of refereeing theory.

\begin{itemize}[label=$\bullet$]
    \item \textbf{Chinese Referee Exams:} We select 118 high-quality questions from the Chinese Football Association (CFA) referee certification exams.
    \item \textbf{International Standards:} We collect 1,100 questions from publicly available referee certification exams conducted by the National Federation of State High School Associations (NFHS) and the Florida High School Athletic Association (FHSAA). To ensure temporal coverage and rule evolution adaptability, the dataset spans multiple years, including 2013, 2017, 2019, 2020, 2021, 2022, 2023, and 2025 for NFHS and 2016 for FHSAA.
\end{itemize}

\subsubsection{Video Data Collection.} For the visual component, we leverage the SoccerNet-MVFoul dataset\cite{VARS}, which provides multi-view video clips of foul incidents. We extract 600 representative samples that feature clear refereeing controversies.

\subsection{Data Curation}\label{Subsec: Data Curation}
To facilitate standardized and objective evaluation, we curate the collected raw data into a unified multiple-choice question format. Each entry in \textbf{SoccerRefBench} consists of a question stem, two to four candidate options, and exactly one correct ground-truth answer. This consistent closed-set QA structure across both textual and video modalities allows for precise accuracy metrics and direct comparison between different models.

For the textual component, data processing is mainly translation. For questions sourced from Chinese Referee Exams, we translated the content into English using LLM-assisted tools, followed by manual verification by domain experts to ensure terminological precision. For the National Federation of State High School Associations (NFHS) and Florida High School Athletic Association (FHSAA) exams, we retained the original official options.

For the video component, we transform the raw annotations from SoccerNet-MVFoul\cite{VARS} into decision-making tasks. To align with the professional referee's disciplinary logic, we map the original ground-truth labels into four distinct severity levels: \textit{No Offence}, \textit{Normal Offence}, \textit{Offence with Yellow Card}, and \textit{Offence with Red Card}. For each video sample, the mapped ground truth serves as the correct option, while the remaining three severity levels are automatically assigned as distractors. This design compels the model to not merely recognize an action, but to adjudicate its severity against strict officiating standards.

\subsection{Knowledge Base Construction}\label{Subsec: Knowledge Base Construction for RAG}
To support accurate decision-making in subsequent intelligent officiating, we construct a specialized domain repository named \textbf{RefKnowledgeDB}. This repository comprises two distinct vector databases designed to assist the automated pipeline in retrieval and reasoning.

\subsubsection{Laws of the Game (LOTG) Knowledge Base ($\mathcal{K}_{rules}$).}We construct the rule-based knowledge base using the latest edition of the IFAB Laws of the Game (2025/26)\cite{IFAB_LOTG_2025}. To preserve structural integrity and ensure precise citation, we parse the raw PDF documents and perform segmentation at the page level. Each page segment is enriched with metadata and encoded into a high-dimensional vector space using OpenAI's embedding model. This page-level granularity ensures that the retrieved context maintains the semantic coherence of specific regulations.
\subsubsection{Classic Case Knowledge Base ($\mathcal{K}_{cases}$).} 
To support high-level Case-Based Reasoning (CBR), we curate a specialized knowledge base comprising 184 historical incidents. These cases are primarily sourced from elite European leagues and the FIFA World Cup, capturing a wide spectrum of standard and controversial officiating scenarios. Each entry is structured in a JSON format that includes a detailed \texttt{Case Description}, the official \texttt{Decision}, and the perceived \texttt{Controversiality} level. This fused text is vectorized to form the retrieval index, while structured attributes are stored as metadata. This design enables the system to retrieve precedents based on semantic similarity across both scenario details and decision outcomes. For a comprehensive breakdown of the data format and source statistics, refer to Appendix~\ref{appendix:kb_detail}.

\section{Methodology}\label{Sec: Method}
We present \textbf{SoccerRef-Agents}, a multi-modal, multi-agent framework designed to mimic the cognitive decision-making process of professional football referees. Unlike traditional end-to-end models, our system decomposes the officiating task into perception, retrieval, legal interpretation, and final adjudication. In this section, we first formulate the problem (Sec.~\ref{Subsec: Problem Formulation}), show our multi-agent architecture(Sec.~\ref{Subsec: Multi-Agent Architecture}) and detail the agentic workflow for both textual and video-based scenarios (Sec.~\ref{Workflow and Reasoning Chains}).
\subsection{Problem Formulation}\label{Subsec: Problem Formulation}
We define the refereeing task as a conditional generation problem that requires both a discrete decision and a legally grounded explanation. Given a query input $q$ consisting of a textual description $q_{txt}$ or in some cases, together with a video clip $q_{vid}$, the system aims to generate a final decision $y$ and a corresponding explanation $e$. The decision $y$ denotes the predicted outcome, and it represents the correct option for theoretical questions or the disciplinary action for video scenarios. Complementing this, $e$ provides a comprehensive justification that synthesizes the specific input evidence with the Laws of the Game\cite{IFAB_LOTG_2025} and historical precedents. Based on the above definitions, the refereeing task can be formally expressed as follows:
$$(y, e) = \mathcal{F}(q; \mathcal{K}_{rules}, \mathcal{K}_{cases}),$$
where $\mathcal{F}$ represents the \textbf{SoccerRef-Agents} system, and $\mathcal{K}$ denotes the external knowledge bases as mentioned in Sec.~\ref{Subsec: Knowledge Base Construction for RAG}.

\subsection{Multi-Agent Architecture}\label{Subsec: Multi-Agent Architecture}
The core of our framework is a collaborative ecosystem comprising four specialized agents and one central decision-maker. Each agent is designed to handle a distinct aspect of the refereeing workflow, from perception to legal reasoning.
To mitigate hallucinations and ensure legally reasoning, the agents are supported by two external vectorized knowledge bases that serve as long-term memory.

\subsubsection{Agents.}

Each agent plays a complementary role, collectively emulating the collaborative decision-making process of a professional refereeing team:

\begin{itemize}[label=$\bullet$]
    
    \item \textbf{Video Agent:} A Vision-Language Model (VLM) specialist responsible for parsing raw video frames into structured textual descriptions and providing initial option recommendations based on visual perception.
    
    \item \textbf{Rule Agent:} Responsible for interpreting the \textit{Laws of the Game}\cite{IFAB_LOTG_2025}. It takes the retrieved rules and summarizes their applicability to the current scenario, ensuring decisions are grounded in the official statute.
    
    \item \textbf{Case Agent:} Specializes in historical analogy. It compares the current situation with retrieved classic cases to provide precedent-based insights, helping to resolve ambiguities through similar past judgments.
    
    \item \textbf{Context Agent:} Extracts and narrates the match importance (e.g., Derby, Final, League match) provided in the metadata text. This contextual information is crucial for judging subjective factors like "intent" and "game management."
    
    \item \textbf{Chief Referee Agent:} The final decision-maker. It aggregates the outputs from all subordinate agents (Video, Rule, Case, and Context) to synthesize a comprehensive rationale and derive the final verdict.
\end{itemize}

\subsubsection{Knowledge Retrieval Mechanism.}
Among the aforementioned agents, the \textit{Rule Agent} and \textit{Case Agent} serve as the most direct and essential components for football rule-based decision-making. To support these two specific agents with precise domain knowledge, we implement a Retrieval-Augmented Generation (RAG) mechanism. This process retrieves the most relevant knowledge segments based on the semantic similarity between the query and the database entries.

To measure the semantic relevance, we utilize cosine similarity. The retrieval process extracts the top-$K$ most relevant segments by maximizing this similarity score. Formally, given a query $q'$ (which can be a text question or a generated video description) and a knowledge base $\mathcal{K}$ (either $\mathcal{K}_{rules}$ or $\mathcal{K}_{cases}$), the retrieval set $\mathcal{R}$ is defined as:

\begin{equation}
\mathcal{R}(q', \mathcal{K}) = \mathop{\text{TopK}}\limits_{k \in \mathcal{K}} \left( \text{Sim}(\mathbf{E}(q'), \mathbf{E}(k)) \right)
\end{equation}

\noindent where $\mathbf{E}(\cdot)$ is the embedding function mapping inputs to a high-dimensional vector space. The similarity function $\text{Sim}(\cdot, \cdot)$ is explicitly calculated as the dot product of the normalized vectors:

\begin{equation}
\text{Sim}(\mathbf{u}, \mathbf{v}) = \frac{\mathbf{u} \cdot \mathbf{v}}{\| \mathbf{u} \| \| \mathbf{v} \|} = \frac{\sum_{i=1}^{d} u_i v_i}{\sqrt{\sum_{i=1}^{d} u_i^2} \sqrt{\sum_{i=1}^{d} v_i^2}}
\end{equation}

\noindent Here, $\mathbf{u} = \mathbf{E}(q')$ and $\mathbf{v} = \mathbf{E}(k)$ represent the embedding vectors of the query and the knowledge segment, respectively. The $\text{TopK}$ operator selects the $K$ segments with the highest similarity scores, which are then passed to the downstream agents for summarization and reasoning.

\subsection{Workflow and Reasoning Chains}\label{Workflow and Reasoning Chains}
\begin{figure*}[t] 
  \centering
  \includegraphics[width=1\textwidth]{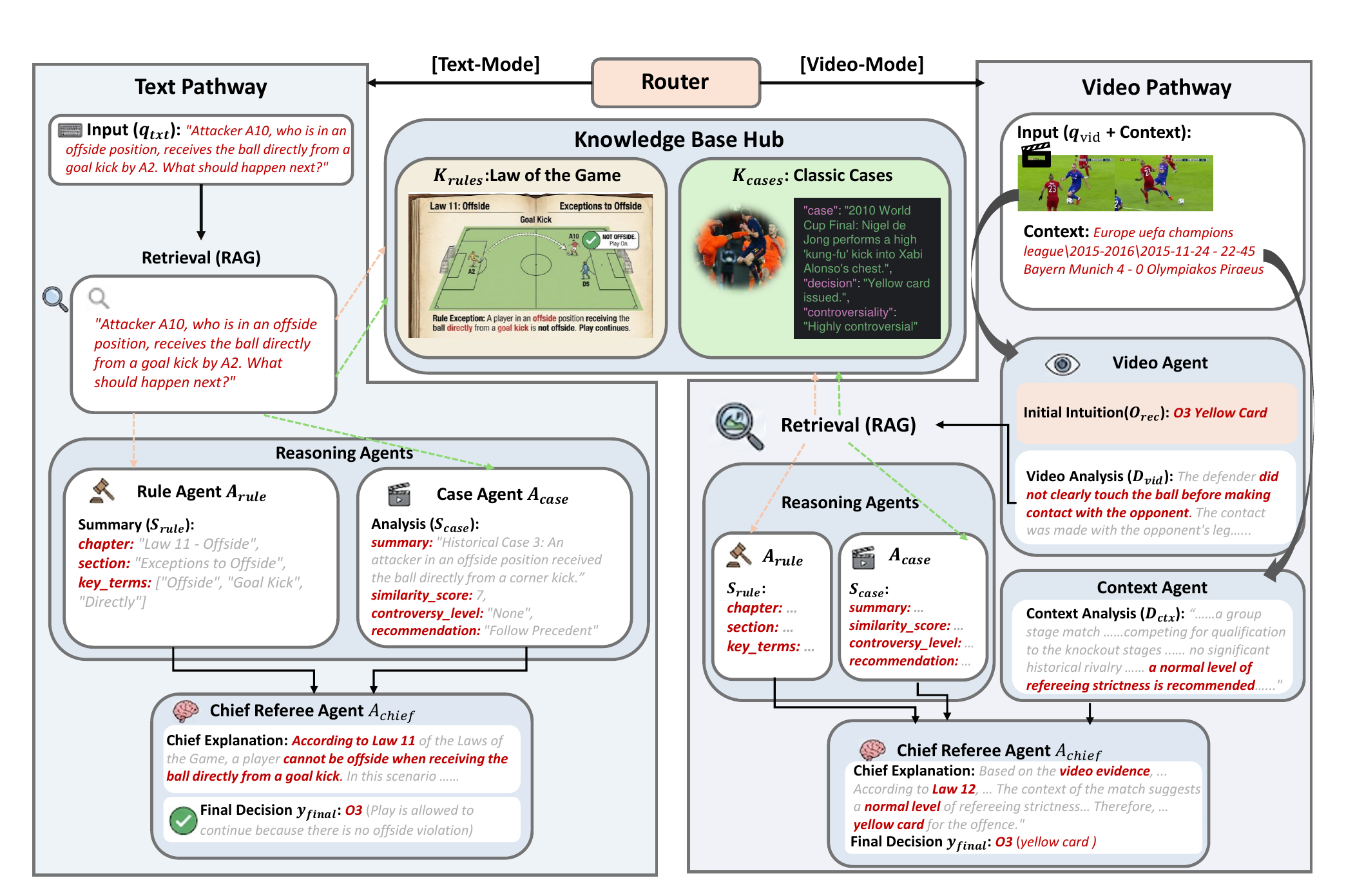}
  
  \caption{Overview of the system's dual-pathway reasoning workflow. Depending on the input modality identified by the Router, the system executes either the Text-Mode or Video-Mode pipeline. The \textbf{Text-Mode Pipeline} directly utilizes the input query for retrieval. The \textbf{Video-Mode Pipeline} features a specialized Cross-Modal RAG mechanism, where the Video Agent converts visual information into a textual Video Analysis ($D_{vid}$) to bridge the semantic gap for retrieving laws and cases. Both pipelines culminate in the Chief Referee Agent, which synthesizes specialist summaries ($S_{rule}, S_{case}$) and comprehensive context to render final adjudication.}
  \label{fig:workflow_diagram}
\end{figure*}

Given the variations in the form and focus of the queries we receive, we build upon previous 5 agents to construct a collaborative multi-agent system, which is designed with two distinct reasoning pipelines based on the input modality.

\subsubsection{Text-Mode Reasoning Pipeline.} When the Router identifies the input as text, the workflow proceeds as follows: (i)\textit{Retrieval:} The input text $q_{txt}$ is used directly as the query to retrieve the top-3 relevant segments from both $\mathcal{K}_{rules}$ and $\mathcal{K}_{cases}$. (ii)\textit{Specialist Analysis:} The \textit{Rule Agent} generates a rule summary ($S_{rule}$) and a concise logic chain linking the text to the rule based on the retrieved laws. Simultaneously, the \textit{Case Agent} generates a precedent summary ($S_{case}$) based on similar historical scenarios. (iii)\textit{Final Adjudication:} The \textit{Chief Referee Agent} receives the original text $q_{txt}$, $S_{rule}$, and $S_{case}$. It synthesizes this legal and historical context to select the correct option and generate an explanation.
$$S_{rule} = \mathcal{A}_{rule}(\mathcal{R}(q_{txt}, \mathcal{K}_{rules}))$$$$S_{case} = \mathcal{A}_{case}(\mathcal{R}(q_{txt}, \mathcal{K}_{cases}))$$$$(y,e) = \mathcal{A}_{chief}(q_{txt} , S_{rule} , S_{case})$$

\subsubsection{Video-Mode Reasoning Pipeline.} When the Router identifies the input as video, the workflow proceeds as follows: (i)\textit{Visual Perception:} The \textit{Video Agent} processes the video $q_{vid}$. It outputs two key components: a preliminary recommendation ($O_{rec}$) and a detailed Video Analysis ($D_{vid}$), which describes the player actions, contact point, and intensity in natural language. (ii)\textit{Contextualization:} The \textit{Context Agent} processes the accompanying context text $q_{ctx}$ to generate a match background description ($D_{ctx}$). (iii)\textit{Cross-Modal Retrieval:} Crucially, we use the generated Video Analysis $D_{vid}$ as the query embedding for the RAG module. This bridges the modality gap, allowing us to retrieve relevant rules and cases based on the visual narrative.The \textit{Rule Agent} produces $S_{rule}$ based on $D_{vid}$ driven retrieval, and the \textit{Case Agent} produces $S_{case}$ based on $D_{vid}$ driven retrieval. (iv)\textit{Aggregation \& Decision:} The \textit{Chief Referee Agent} synthesizes the comprehensive information set: $\{O_{rec}, D_{vid}, D_{ctx}, S_{rule}, S_{case}\}$. This allows the Chief Referee to validate the \textit{Video Agent}'s initial intuition against strict rules and precedents before making the final decision.

$$D_{vid}, O_{rec} = \mathcal{A}_{video}(q_{vid})$$$$D_{ctx} = \mathcal{A}_{context}(q_{ctx})$$$$S_{rule} = \mathcal{A}_{rule}(\mathcal{R}(D_{vid}, \mathcal{K}_{rules}))$$$$S_{case} = \mathcal{A}_{case}(\mathcal{R}(D_{vid}, \mathcal{K}_{cases}))$$$$(y,e) = \mathcal{A}_{chief}(O_{rec} , D_{vid} , D_{ctx} , S_{rule} , S_{case})$$

\section{Experiments}\label{Sec: Experiments}

Based on the multi-agent system we have constructed, we conducted extensive experiments to evaluate its performance. This section outlines the experimental protocols in Sec.~\ref{subsec: Experimental Settings}, reports comprehensive quantitative comparisons against state-of-the-art MLLMs in Sec.~\ref{subsec: Quantitative Results} and Sec.~\ref{subsec: Ablation Studies}, shows human evaluation in Sec.~\ref{subsec: Human Evaluation}, and concludes with qualitative analysis in Sec.~\ref{subsec: Qualitative Analysis}.

\subsection{Experimental Settings}\label{subsec: Experimental Settings}

\subsubsection{Baselines.}
We evaluate our \textbf{SoccerRef-Agents} against a comprehensive suite of state-of-the-art MLLMs on \textbf{SoccerRefBench}, covering both proprietary commercial APIs and open-source models. \textit{(i) Commercial APIs:} We leading commercial APIs with top-tier performance on broad multimodal and reasoning tasks, including GPT-4o\cite{gpt4o}, Claude 4.5 Sonnet\cite{claude45} and Gemini 2.5 Flash\cite{gemini25}. These API models are prompted with detailed system instructions but rely solely on their internal parametric knowledge (i.e., without external retrieval). \textit{(ii) Open-Source Models:} We evaluate the Qwen3-VL series (8B and 32B)\cite{qwen3vl} as competitive public VLMs for both text and video understanding. Additionally, for the text-only task, we include DeepSeek-V3\cite{deepseek} as a strong textual baseline.

\subsubsection{Metrics.}
As for the evaluation on our proposed benchmark \textbf{SoccerRefBench}, we employ Accuracy (Acc) as the primary quantitative metric for both subsets. 
To further assess the interpretability and practical validity of the system's decisions, we introduce a human evaluation protocol. In this protocol, evaluations are conducted by individuals with professional refereeing experience. The details of this protocol are elaborated in Section \ref{subsec: Human Evaluation}.

\subsection{Quantitative Results}\label{subsec: Quantitative Results}

We present the comparative results on \textbf{SoccerRefBench} in Table \ref{tab:main_results}.Our observations are as follows: (i) The results on \textbf{SoccerRefBench} exhibit a broad performance spectrum, effectively differentiating the capabilities of various MLLMs. Accuracy scores span from 46.88\% to 79.56\% in the text domain and 23.50\% to 40.17\% in the video domain. This significant variance underscores the complexity and diversity of our benchmark, confirming that it serves as a rigorous testing ground capable of distinguishing the performance gaps between standard open-source models, advanced commercial APIs, and specialized agentic systems. (ii) As shown in Table \ref{tab:main_results}, \textbf{SoccerRef-Agents} achieves the highest accuracy across both modalities. In the text domain, our system reaches \textbf{79.56\%}, surpassing the strongest baseline (GPT-4o\cite{gpt4o}) by 1.73\%. This indicates that even highly capable generalist models like GPT-4o struggle with the professional soccer referee theories and highly specific knowledge required for professional referee exams, whereas our RAG-enhanced \textit{Rule Agent} effectively closes this gap. (iii) The video domain task proves to be extremely challenging for all models, with no baseline exceeding 38\% accuracy. This highlights the intrinsic difficulty of distinguishing subtle foul grades solely from broadcast footage. Despite this, \textbf{SoccerRef-Agents} achieves state-of-the-art performance with \textbf{40.17\%} accuracy, outperforming GPT-4o\cite{gpt4o} (+2.5\%) and significantly surpassing open-source video models like Qwen3-VL\cite{qwen3vl} (+15.8\%). This performance gain validates our cross-modal design, where the system aligns visual descriptions with textual rules and historical knowledge to make more informed decisions.

\begin{table}[t]
    \centering
    \caption{Main Results on \textbf{SoccerRefBench}. We compare \textbf{SoccerRef-Agents} with state-of-the-art commercial and open-source MLLMs. The best performance is highlighted in \textbf{bold}, and the second best is \underline{underlined}. Note that DeepSeek-V3\cite{deepseek} is a text-only model and is thus evaluated only on the Text subset.}
    \label{tab:main_results}
    \resizebox{\linewidth}{!}{
    \begin{tabular}{l | >{\centering\arraybackslash}p{2.3cm} | >{\centering\arraybackslash}p{2.3cm} | >{\centering\arraybackslash}p{2.3cm}}
        \toprule
        \textbf{Model} & \textbf{Text}(\%) & \textbf{Video}(\%) & \textbf{Overall}(\%) \\
        \midrule
        
        \multicolumn{4}{c}{\textit{Open-Source Models}} \\
        \midrule
        Qwen3-VL-8B\cite{qwen3vl} & 46.88 & 23.50 & 39.16 \\
        Qwen3-VL-32B\cite{qwen3vl} & 56.57 & 24.33 & 45.93 \\
        DeepSeek-V3\cite{deepseek} & 65.35 & - & - \\
        \midrule
        
        \multicolumn{4}{c}{\textit{Commercial APIs}} \\
        \midrule
        Gemini 2.5 Flash\cite{gemini25} & 69.38 & 26.33 & 55.17 \\
        Claude 4.5 Sonnet\cite{claude45} & 65.19 & 34.67 & 55.12 \\
        GPT-4o\cite{gpt4o} & \underline{77.83} & \underline{37.67} & \underline{64.58} \\
        \midrule
        
        \textbf{SoccerRef-Agents (Ours)} & \textbf{79.56} & \textbf{40.17} & \textbf{66.56} \\
        \bottomrule
    \end{tabular}
    }
\end{table}

\subsection{Ablation Studies}\label{subsec: Ablation Studies}

To further investigate the individual contributions of our specialized knowledge bases and the synergy between the \textit{Rule Agent} and\textit{ Case Agent}, we conduct ablation experiments on the \textbf{SoccerRefBench}. We evaluate three configurations: (i) \textbf{SoccerRef-Agents (Full)}, our complete multi-agent framework; (ii) \textbf{w/o Rule Agent}, where the system relies solely on historical precedents and internal knowledge; and (iii) \textbf{w/o Case Agent}, where the system only retrieves information from the \textit{Laws of the Game}\cite{IFAB_LOTG_2025}.

\begin{table}[h]
    \centering
    \small
    \caption{Ablation Study of Knowledge Components. We report the Accuracy (\%) on the Text (1,218 Qs), Video (600 Qs) subsets, and the weighted Overall performance.}
    \label{tab:ablation_results}
    \begin{tabular}{l | >{\centering\arraybackslash}p{2.3cm} | >{\centering\arraybackslash}p{2.3cm} | >{\centering\arraybackslash}p{2.3cm}}
        \toprule
        \textbf{Variant} & \textbf{Text} & \textbf{Video} & \textbf{Overall} \\
        \midrule
        w/o Rule Agent & 78.90\% & \textbf{42.50\%} & \textbf{66.89\%} \\
        w/o Case Agent & \textbf{79.89\%} & 39.17\% & 66.45\% \\
        \midrule
        \textbf{SoccerRef-Agents (Full)} & 79.56\% & 40.17\% & 66.56\% \\
        \bottomrule
    \end{tabular}
\end{table}

The results in Table \ref{tab:ablation_results} reveal a notable divergence in knowledge utilization across modalities. The text task exhibits rule-heavy characteristics, achieving peak accuracy (\textbf{79.89\%}) when relying solely on the \textit{Rule Agent}; here, historical cases act as noise that interferes with strictly literal rule interpretations. Conversely, the video task is inherently case-heavy, with the \textit{w/o Rule Agent} variant reaching the highest accuracy (\textbf{42.50\%}). This suggests that practical officiating requires the guide of precedents to navigate subjective visual scenarios where rigid rule-following may prove too restrictive. Although individual components yield marginal gains in specialized domains, our full \textbf{SoccerRef-Agents} framework provides the most robust balance across both modalities. Crucially, as shown in our human evaluation, integrating both the rule guideline and historical precedents is essential for generating the professionally justifiable and legally grounded explanations required for multi-modal refereeing.

\subsection{Human Evaluation of Explanation Quality}\label{subsec: Human Evaluation}

Accuracy alone does not fully capture the utility of an AI referee; the reasoning process must be transparent, authoritative, and legally binding. Since current automatic metrics correlate poorly with logical validity in this specialized domain, we conduct a rigorous human evaluation campaign.

We focus our evaluation on comparing \textbf{SoccerRef-Agents} against the top-performing baseline, \textbf{GPT-4o}\cite{gpt4o}. We randomly sample 100 textual questions and 50 video questions from our \textbf{SoccerRefBench}. We invite three certified referees authorized by the Chinese Football Association to act as expert annotators. To ensure objectivity, the evaluation is conducted in a blind setting where the source of each explanation is masked.Unlike general readability scores, our scoring rubric is strictly tailored to professional officiating standards. Experts rate each explanation on a 1-5 Likert scale whose definition can be checked in the appendix~\ref{appendix:human_evaluation}.

The comparative evaluation results from three expert referees are detailed in Table \ref{tab:human_eval}. While GPT-4o\cite{gpt4o} exhibits strong linguistic fluency, experts noted it frequently falters in rule adherence, inventing plausible-sounding but non-existent regulations. In contrast, our system effectively minimizes such domain-specific hallucinations. By explicitly citing the \textit{Laws of the Game}\cite{IFAB_LOTG_2025} and anchoring reasoning in historical precedents, \textbf{SoccerRef-Agents} provides more trustworthy and legally grounded explanations, closely mirroring professional cognitive workflows.

\begin{table}[t]
    \centering
    \caption{Human Evaluation Results. Three professional referees rated the explanation quality (1-5 scale). Columns denote average scores for Text (100 Qs), Video (50 Qs), and Overall weighted performance.}
    \label{tab:human_eval}
    \resizebox{\linewidth}{!}{
    \begin{tabular}{l|ccc|ccc|ccc|ccc}
        \toprule
        \multirow{2}{*}{\textbf{Model}} & \multicolumn{3}{c|}{\textbf{Referee 1}} & \multicolumn{3}{c|}{\textbf{Referee 2}} & \multicolumn{3}{c|}{\textbf{Referee 3}} & \multicolumn{3}{c}{\textbf{Average}} \\
         & \textbf{Text} & \textbf{Video} & \textbf{Overall} & \textbf{Text} & \textbf{Video} & \textbf{Overall} & \textbf{Text} & \textbf{Video} & \textbf{Overall} & \textbf{Text} & \textbf{Video} & \textbf{Overall} \\
        \midrule
        GPT-4o\cite{gpt4o} & 3.63 & 2.70 & 3.32 & 3.65 & 2.70 & 3.33 & 3.72 & 2.80 & 3.41 & 3.67 & 2.73 & 3.36 \\
        \textbf{Ours} & \textbf{3.76} & \textbf{2.76} & \textbf{3.43} & \textbf{4.09} & \textbf{3.24} & \textbf{3.81} & \textbf{3.96} & \textbf{3.18} & \textbf{3.70} & \textbf{3.94} & \textbf{3.06} & \textbf{3.65} \\
        \bottomrule
    \end{tabular}
    }
\end{table}

\subsection{Qualitative Analysis}\label{subsec: Qualitative Analysis}

\begin{figure}[t]
    \centering
    \includegraphics[width=1.0\linewidth]{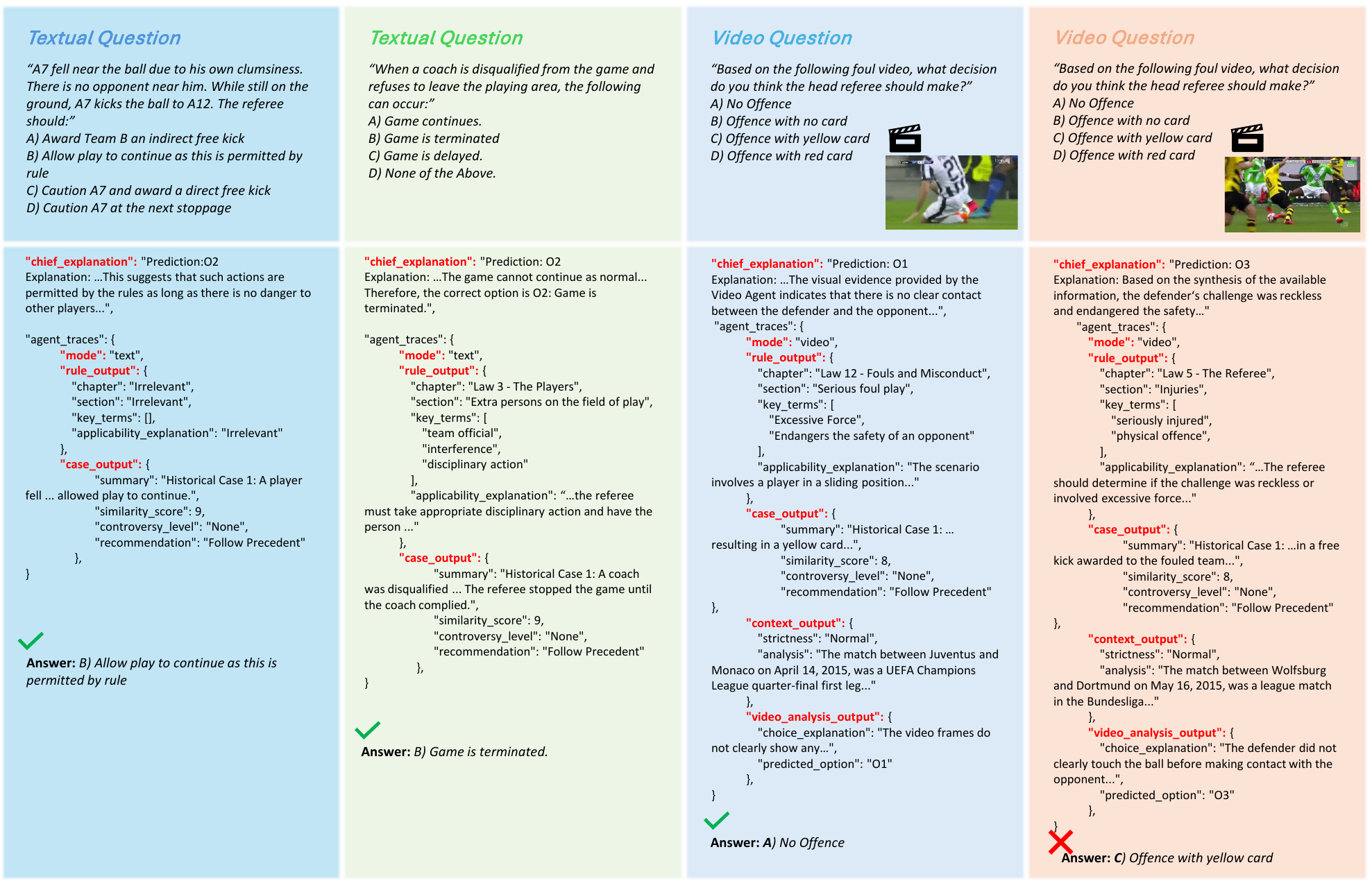}
    \caption{Qualitative examples of \textbf{SoccerRef-Agents} on \textbf{SoccerRefBench}. The figure illustrates the step-by-step reasoning chains for both textual (left two columns) and video-based (right two columns) queries. Key intermediate outputs from the Rule Agent, Case Agent, and Video Agent are explicitly logged in \texttt{agent\_traces}, providing legally grounded justifications for the final decision.}
    \label{fig:qualitative_results}
\end{figure}

To demonstrate the interpretability and reasoning depth of \textbf{SoccerRef-Agents}, we present a visualization of the decision-making process in Figure \ref{fig:qualitative_results}. Unlike black-box models that output a label without context, our system provides a transparent "glass-box" view of its logic through detailed \texttt{agent\_traces}.

As illustrated in Figure \ref{fig:qualitative_results}, our framework effectively handles diverse modalities. For theoretical questions, the system accurately retrieves specific chapters from the \textit{Laws of the Game}. For instance, in the second column regarding a disqualified coach, the \textit{Rule Agent} correctly pinpoints "Law 3 - The Players" and "Extra persons on the field of play". Simultaneously, the \textit{Case Agent} retrieves a historical precedent with a high similarity score (9/10), reinforcing the \textit{Chief Referee Agent}'s decision to "terminate the game" rather than merely delaying it. In complex visual scenarios, the \textit{Video Agent} acts as the primary perceiver. In the third column, the agent explicitly notes that "the video frames do not clearly show any [contact]". This visual evidence drives the \textit{Chief Referee Agent} to conclude "No Offence," aligning with the ground truth. Also, in the video questions, the \textit{Context Agent} identifies the match such as "league match in the Bundesliga between Wolfsburg and Dortmund". These information provides match context for the \textit{Chief Referee Agent} to confirm the standard of judgment. This granular visibility into the reasoning process confirms that \textbf{SoccerRef-Agents} does not merely memorize answers but actively constructs a legal argument, significantly enhancing trust in automated officiating. More results are provided in the Appendix~\ref{appendix:qualitative}.

\section{Conclusion}
This paper presents a comprehensive framework for explainable and standardized automated soccer refereeing for multimodal scenarios. Concretely, we introduce \textbf{SoccerRefBench}, a unique multimodal benchmark bridging theoretical exams and practical video judgments, accompanied by a specialized vector knowledge base \textbf{RefKnowledgeDB} to ensure legal precision. To tackle the complexity of officiating logic, we establish \textbf{SoccerRef-Agents}, a novel multi-agent system that mimics the collaborative workflow of professional referee teams. By leveraging a cross-modal Retrieval-Augmented Generation (RAG) mechanism, our system effectively translates visual evidence into rule-based reasoning, bridging the gap between perception and adjudication. Extensive evaluations have demonstrated the superiority of our framework, showing that it not only achieves higher accuracy than general-purpose models but also provides transparent, citation-backed explanations. We believe this work establishes a new foundation for fair, transparent, and interpretable AI assistants in sports officiating.
%
%
%
\clearpage
\bibliographystyle{splncs04}
\bibliography{references}
\newpage
\appendix

\section{SoccerRefBench Dataset Details}\label{appendix:dataset_detail}

To ensure seamless multimodal evaluation, both textual and video-based queries in \textbf{SoccerRefBench} follow a standardized JSON schema shown in Table \ref{tab:json_schema}. The primary difference lies in the \texttt{materials} field, which for video samples includes the local file path and match context metadata. Listing \ref{lst:soccerref_example} provides examples of the video judgment task and textual task.

\begin{table}
\centering
\caption{Internal JSON Schema of the \textbf{SoccerRefBench} Dataset.}
\label{tab:json_schema}
\begin{tabular}{|l|l|l|}
\hline
Field & Type & Description \\
\hline
Q & string & The question stem (referee theory question) \\
materials & list[string] & Optional auxiliary reference materials \\
O1 & string & Option 1 text \\
O2 & string & Option 2 text \\
O3 & string & Option 3 text \\
O4 & string & Option 4 text \\
openA & string & Ground-truth answer in natural language \\
closeA & string & Ground-truth option label \\
\hline
\end{tabular}
\end{table}

\begin{lstlisting}[language=json, caption={SoccerRefBench Dataset Example}, label=lst:soccerref_example]
{
    "Q": "Based on the following foul video, what decision do you think the head referee should make?",
    "materials": [
      {
        "path": "Dataset/video/SoccerNet/mvfouls/train/action_620/clip_1.mp4",
        "context": "europe_uefa-champions-league\\2014-2015\\2015-04-14 - 21-45 Juventus 1 - 0 Monaco"
      }
    ],
    "openA": "Offence with no card",
    "closeA": "O2",
    "O1": "No offence",
    "O2": "Offence with no card",
    "O3": "Offence with yellow card",
    "O4": "Offence with possible red card"
},
{
    "Q": "Player A1 kicks off to start the second half of the game. Player A1's kick goes directly into Team B's goal. The referee should:",
    "materials": [
      "none"
    ],
    "openA": "Award the goal and restart the match with a kickoff for Team B.",
    "closeA": "O4",
    "O1": "Disallow the goal and have Team A retake the kickoff.",
    "O2": "Disallow the goal and have Team A take an indirect free kick from the halfway line.",
    "O3": "Disallow the goal and award Team B a goal kick.",
    "O4": "Award the goal and restart the match with a kickoff for Team B."
}
\end{lstlisting}

\section{Classic Case Knowledge Base Details}
\label{appendix:kb_detail}

\subsection{Raw Case Format}
Before vectorization into the \textbf{RefKnowledgeDB}, historical cases are curated in a structured format capturing the incident, the official decision, and the perceived controversiality. Listing \ref{list:case_format} showcases the raw entries.

\begin{lstlisting}[language=json, caption={Structured format of historical cases in the Classic Case Knowledge Base.}, label=list:case_format]
{
  "id": 4,
  "case": "2024 Premier League: Declan Rice receives a second yellow card for slightly kicking the ball away to delay the restart.",
  "decision": "Second yellow card and red card issued.",
  "controversiality": "Highly controversial"
},
{
    "id": 61,
    "case": "2012 UCL: John Terry knees Alexis Sanchez in the back during an off-the-ball incident.",
    "decision": "Red card issued.",
    "controversiality": "Non-controversial"
},
{
    "id": 179,
    "case": "2021 La Liga: Referee calls players back from the locker room to take a penalty after VAR review.",
    "decision": "Penalty awarded.",
    "controversiality": "Somewhat controversial"
}
\end{lstlisting}

\subsection{Source Statistics}
The cases are aggregated from authoritative tournament archives and officiating review panels. Table \ref{tab:case_sources} summarizes the distribution of historical precedents.

\begin{table}[h]
    \centering
    \small
    \caption{Distribution of sources for the Classic Case Database.}
    \label{tab:case_sources}
    \begin{tabular}{lc}
        \toprule
        \textbf{Source Authority} & \textbf{Number of Cases} \\
        \midrule
        FIFA World Cup & 72 \\
        Premier League & 40 \\
        UEFA Champions League & 24 \\
        Bundesliga & 19 \\
        La Liga & 17 \\
        Euro Cup & 12 \\
        \bottomrule
    \end{tabular}
\end{table}

\section{Multi-Agent System Prompts}
\label{appendix:prompts}
In this section, we provide the specialized system prompts for each agent in \textbf{SoccerRef-Agents}.

\begin{prompt}{Rule Agent System Prompt}
You are an expert AI Legal Analyst specializing in the IFAB Laws of the Game. Your task is to strictly analyze the provided rule excerpts and identify the exact clause that governs the user's scenario.
\end{prompt}
\begin{prompt}{Rule Agent User Prompt}
    \textbf{Context (Retrieved IFAB Laws):} \texttt{\{retrieved\_rules\}} \\
    \textbf{Scenario:} \textit{``\{query\_text\}''} \\
    
    \textbf{Instructions:}
    \begin{itemize}[leftmargin=1.5em, nosep]
        \item Analyze the specific Law, Section, and Bullet Point.
        \item Prioritize specific offenses (e.g., \textbf{Serious Foul Play}).
        \item Extract text verbatim if relevant.
    \end{itemize}
    
    \textbf{Expected Output:} JSON format with \texttt{direct\_quote}, \texttt{key\_terminology\_match}, and \texttt{confidence} fields.
\end{prompt}
\begin{prompt}{Context Agent User Prompt}
    \textbf{Match Context:} \texttt{\{context\_str\}} \\

    \textbf{Instructions:}
    \begin{itemize}[leftmargin=1.5em, nosep]
        \item Analyze the match importance (e.g., Derby, Final, League match), home/away factors, and potential team rivalries.
        \item Determine the recommended refereeing strictness (Lenient, Normal, Strict).
    \end{itemize}

    \textbf{Expected Output (JSON):} \\
    A JSON object containing:
    \begin{itemize}[leftmargin=1.5em, nosep]
        \item \texttt{strictness}: Recommended enforcement level.
        \item \texttt{analysis}: Brief justification based on match atmosphere and stakes.
    \end{itemize}
\end{prompt}
\begin{prompt}{Video Agent System Prompt}
    You are a professional AI Soccer Referee Assistant. 
    The input contains video frames from a \textbf{live soccer match broadcast replay}. 
    Output JSON only.
\end{prompt}
\begin{prompt}{Video Agent User Prompt}
    \textbf{Input Type:} Broadcast Replay Video \\
    
    \textbf{Question:} \texttt{\{question\_text\}} \\
    
    \textbf{Options:} \\
    \texttt{\{options\_str\}} \\

    \textbf{Instructions:}
    \begin{itemize}[leftmargin=1.5em, nosep]
        \item Analyze the input video carefully.
        \item Select the ONE correct option ID.
        \item Provide a brief explanation in English.
    \end{itemize}

    \textbf{Expected Output (JSON ONLY):} \\
    \{ \\
    \hspace*{1em} \texttt{"choice\_explanation": "...",} \\
    \hspace*{1em} \texttt{"predicted\_option": "O1"} \\
    \}
\end{prompt}
\begin{prompt}{Chief Referee Agent System Prompt}
    You are the Chief Referee Agent, the final decision-maker in a multi-agent soccer refereeing system. 
    Your role is to synthesize evidence from specialized subordinate agents (Rule, Case, Context, and Video) to provide a definitive ruling on complex foul scenarios.
\end{prompt}

\begin{prompt}{Chief Referee Agent User Prompt}
    \textbf{=== QUESTION DATA ===} \\
    \textbf{Question:} \texttt{\{question\_text\}} \\
    \textbf{Options:} \texttt{\{options\_text\}} \\
    \textit{PS: 'A\#' means player of team A with jersey number \#, same for 'B\#'.} \\

    \textbf{=== SUBORDINATE AGENT REPORTS ===} \\
    \textbf{[1] Reference Law:} \\
    \texttt{\{rule\_str\_placeholder\}} \textit{\%(Includes Text of Law, Match Logic, and Confidence)\%} \\

    \textbf{[2] Reference Precedents:} \\
    \texttt{\{case\_str\_placeholder\}} \textit{\%(Includes Valid Precedent or No Precedent status)\%} \\

    \textbf{[3] Reference Context (Video Mode Only):} \\
    \texttt{\{context\_analysis\}} \\

    \textbf{[4] Visual Evidence (Video Agent):} \\
    \begin{itemize}[leftmargin=1.5em, nosep]
        \item \textbf{Video Agent's Choice Explanation:} \texttt{\{desc\}}
        \item \textbf{Video Agent's Initial Intuition:} \texttt{\{pred\}}
    \end{itemize}

    \textbf{=== INSTRUCTIONS ===}
    \begin{itemize}[leftmargin=1.5em, nosep]
        \item Analyze the provided input text and subordinate reports carefully.
        \item Select the most correct ONE option ID.
        \item Provide a brief explanation in English.
    \end{itemize}

    \textbf{OUTPUT FORMAT:} \\
    \texttt{Prediction: [Option ID]} \\
    \texttt{Explanation: [Reasoning]}
\end{prompt}
\section{Details on human evaluation of explanation quality}\label{appendix:human_evaluation}
We created an html webpage for the professional referees to score the quality of explanation output by the model
\subsection{The definition of Likert Scale}
\begin{itemize}[label=$\bullet$]
    \item \textbf{5 - Perfect:} Precisely identifies the foul or answers the theory question with correct legal citations. The causal logic is clear, and the terminology is professional.
    \item \textbf{4 - Good:} The verdict and key descriptions are correct. May contain minor terminological imprecisions or verbose explanations, but the core reasoning is valid.
    \item \textbf{3 - Fair:} The conclusion is correct, but the explanation contains slight hallucinations or misses rule support.
    \item \textbf{2 - Poor:} Cites incorrect rules or describes actions significantly inconsistent with the video evidence.
    \item \textbf{1 - Nonsense:} Contains severe hallucinations or chaotic logic that contradicts basic football common sense.
\end{itemize}
\subsection{Human Evaluation Interface}
\begin{figure}[h]
    \centering
    \includegraphics[width=1.0\linewidth]{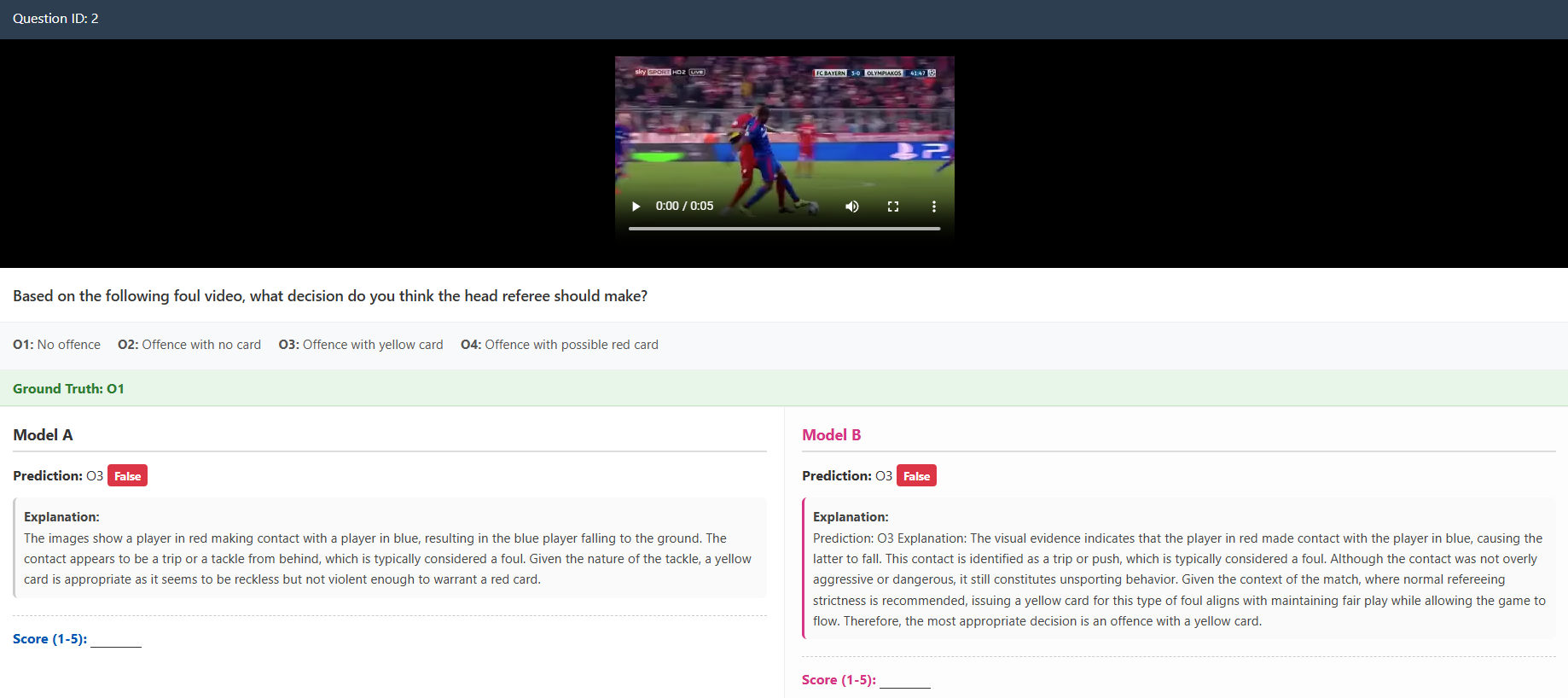}
    \caption{Human Evaluation Interface}
    \label{fig:interface1}
\end{figure}

\begin{figure}[h]
    \centering
    \includegraphics[width=1.0\linewidth]{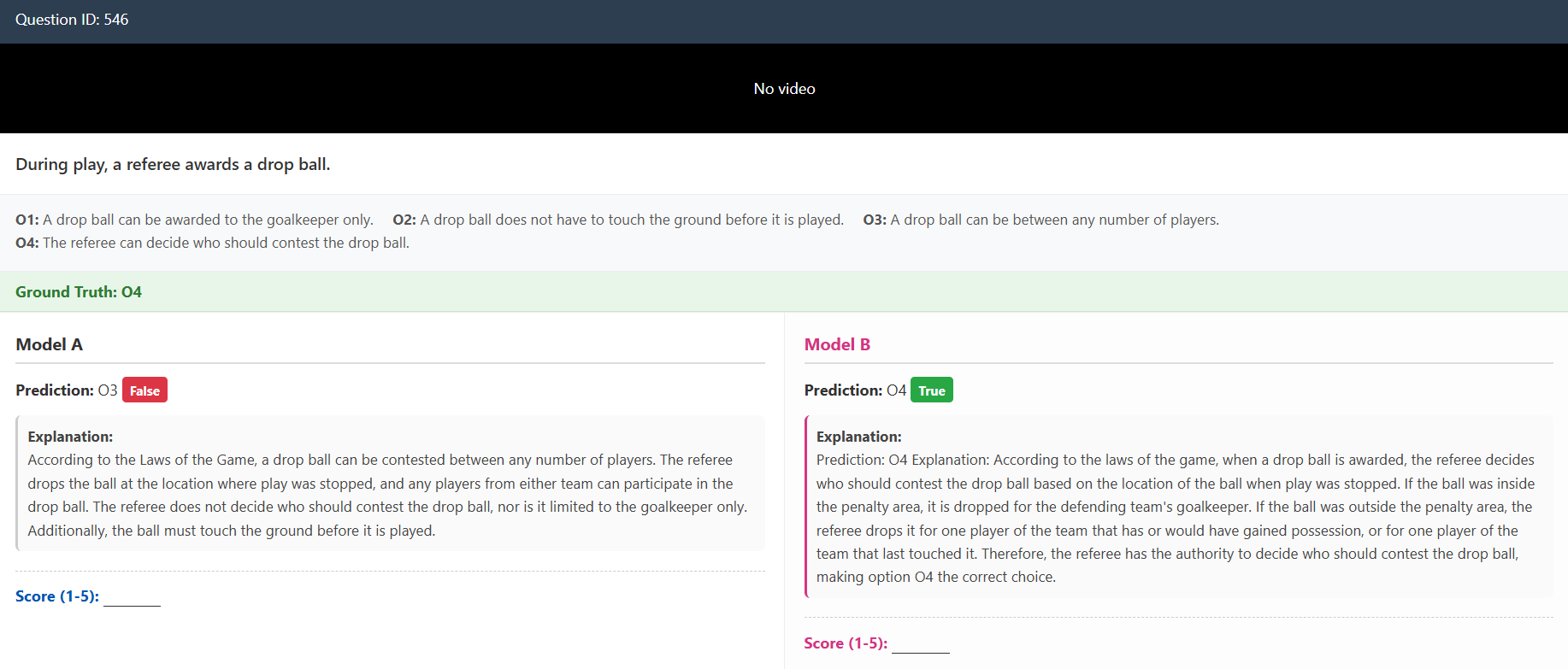}
    \caption{Human Evaluation Interface}
    \label{fig:interface2}
\end{figure}

\section{Additional Qualitative Results}
\label{appendix:qualitative}
\begin{figure}[h]
    \centering
    \includegraphics[width=1.0\linewidth]{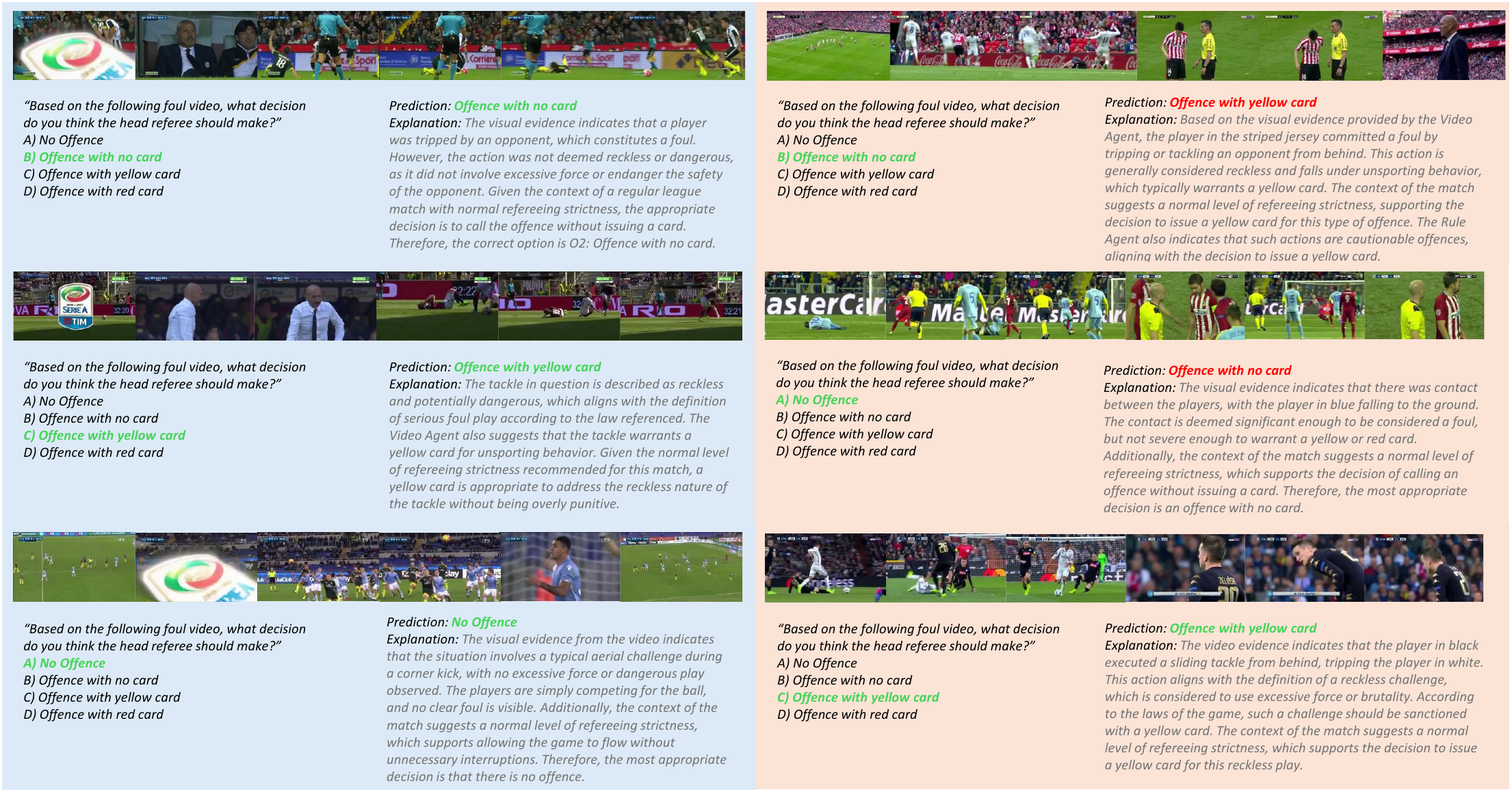}
    \caption{Extended qualitative results}
    \label{fig:more_qualitative1}
\end{figure}

\begin{figure}[h]
    \centering
    \includegraphics[width=1.0\linewidth]{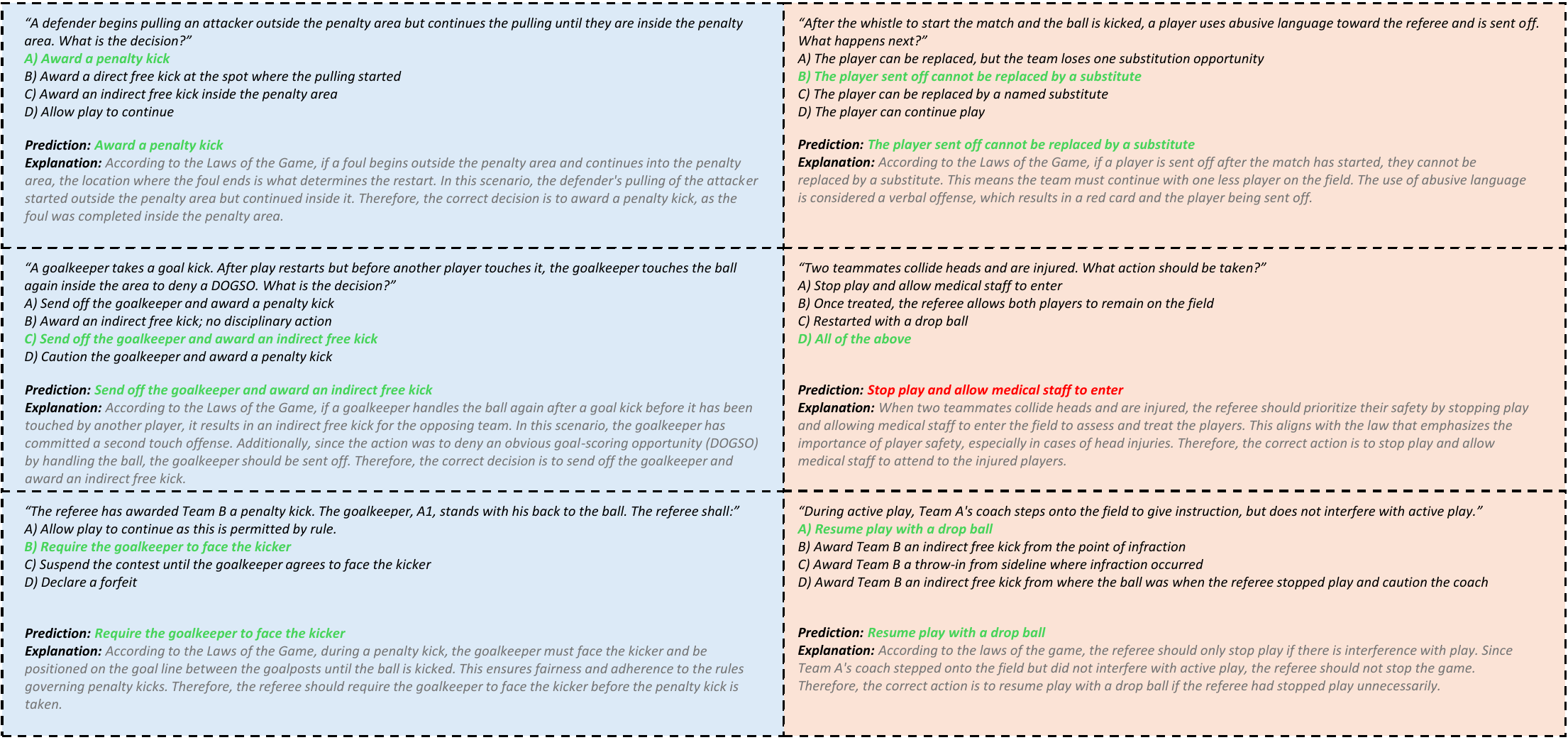}
    \caption{Extended qualitative results}
    \label{fig:more_qualitative2}
\end{figure}

Figure \ref{fig:more_qualitative1} and Figure \ref{fig:more_qualitative2} present further examples of \textbf{SoccerRef-Agents}'s performance on \textbf{SoccerRefBench}.


\end{document}